\documentclass{elsarticle}

\usepackage{graphicx} 
\usepackage{amsmath}
\usepackage{mathtools}
\usepackage{array}
\usepackage{tabularx}

\begin{document}
\begin{frontmatter}

\title{Theater Aid System for the Visually Impaired Through Transfer Learning of Spatio-Temporal Graph Convolution Networks}

\author[1]{Leyla Benhamida\corref{cor1}%
}
\ead{lbenhamida@usthb.dz}
\author[1]{Slimane Larabi}
\ead{slarabi@usthb.dz}
\affiliation[1]{organization={RIIMA Laboratory, Computer Science Faculty, USTHB University},
addressline={BP 32 EL ALIA},
postcode={16111},
city={Algiers},
country={Algeria}
}

\let\WriteBookmarks\relax
\def\floatpagepagefraction{1}
\def\textpagefraction{.001}

\begin{abstract}
The aim of this research is to recognize human actions performed on stage to aid visually impaired and blind individuals. To achieve this, we have created a theatre human action recognition system that uses skeleton data captured by depth image as input. We collected new samples of human actions in a theatre environment, and then tested the transfer learning technique with three pre-trained Spatio-Temporal Graph Convolution Networks for skeleton-based human action recognition: the spatio-temporal graph convolution network, the two-stream adaptive graph convolution network, and the multi-scale disentangled unified graph convolution network. We selected the NTU-RGBD human action benchmark as the source domain and used our collected dataset as the target domain. We analyzed the transferability of the pre-trained models and proposed two configurations to apply and adapt the transfer learning technique to the diversity between the source and target domains. The use of transfer learning helped to improve the performance of the human action system within the context of theatre. The results indicate that Spatio-Temporal Graph Convolution Networks is positively transferred, and there was an improvement in performance compared to the baseline without transfer learning.
\end{abstract}
\end{frontmatter}

\section{Introduction}

\begin{figure}[ht]
{
    \centering
    \includegraphics[scale=1]{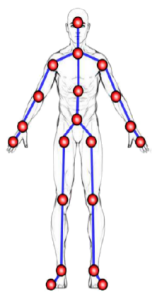}
    \caption{Skeleton data: red points represent the provided human body key joints by kinect v1.}
    \label{kinectV1_joints}}
\end{figure}

Numerous aid systems have been developed to assist people with visual impairments in their daily tasks, including understanding the scene from depth images \cite{Zatout2021} \cite{Zatout2019} \cite{Zatout2020}, identifying objects \cite{34}, navigating their environment and avoiding obstacles \cite{31,32}, visual positioning from depth images \cite{Ibel2022} \cite{Ibel2020} \cite{Ibel2020_2}, image captioning for visually impaired \cite{Delloul2022_2}
\cite{Delloul2022} and  Human Action Recognition and Coding based on Skeleton \cite{Benhamida2022}. Many of these systems rely on computer vision methods, such as smart glasses with built-in cameras that analyze the user's surroundings and provide audio descriptions or directions based on the captured images \cite{30}. Some systems even use facial recognition technology to help users recognize people they know \cite{33}. Despite the various aid systems available to accomplish daily activities, there is a dearth of systems that provide entertainment options for visually impaired individuals. Theaters and cinemas, for example, are not accessible to them as descriptions of the actions and movements of the actors are necessary for proper understanding. To our knowledge, no existing works provide descriptions of human actions on stage for the visually impaired. Therefore, the primary objective of our research is to create a system that can recognize human actions on stage, allowing visually impaired individuals to enjoy theaters and similar venues.\\
In this research paper, we present a comprehensive framework for recognizing human actions in theatrical scenes using computer vision and deep learning models. Our framework leverages input data captured by the Microsoft Kinect v1 sensor, specifically the actor's actions performed on stage. The Kinect sensor provides three types of data: RGB, depth, and skeleton data. In this study, we focus on utilizing skeleton data, which represents a collection of 3D positions of human body joints (Figure.\ref{kinectV1_joints}), to develop an effective approach for human action recognition.

One widely-used deep learning model for skeleton-based Human Action Recognition (HAR) is the Spatio-Temporal Graph Convolution Network (ST-GCN), known for its success in challenging benchmarks such as NTU-RGBD \cite{1,2}. Our goal is to explore the applicability of ST-GCN and similar models for action recognition in theatre scenes by employing transfer learning techniques to enhance their performance. To accomplish this, we curate a new dataset of human action sequences specifically recorded in a theatre environment.\\
Through this research, we investigate and analyze the transferability of deep learning models for skeleton-based HAR in the context of theatre actions—a topic that has not been previously explored in existing literature.\\
The paper is structured as follows: In Section 2, we commence with an extensive review of the skeleton-based human action recognition approach utilizing Spatio-Temporal Graph Convolution Network (GCN) models. This review encompasses in-depth explanations of ST-GCN, 2s-AGCN, and MS-G3D, providing a comprehensive understanding of their functionalities. Subsequently, we delve into the concept of transfer learning, offering a precise definition and discussing its relevance to our research.
Moving forward to Section 3, we elucidate the objectives pursued in this study and provide a well-founded justification for the choices made during the course of our research. Section 4 entails a detailed description of the dataset we have collected, and we introduce our proposed method for human action recognition.
Lastly, in Section 5, we present the experimental setup conducted to evaluate the effectiveness of our approach, along with an in-depth discussion of the results obtained. Finally, section 6, summarizes the key findings and the contributions of our research.\\
\section{Related Works}

Human action recognition (HAR) is a complex research field that finds application in numerous domains. In the past years, various techniques were developed, with the majority relying on computer vision approaches with video sequences captured by camera sensors. One of the most frequently used sensors in HAR systems today is the Microsoft Kinect. It captures both RGB and depth information, providing a richer representation of the scene than traditional RGB cameras. The depth information can be used to extract features, such as joint positions and orientations, which helps boost the performance of the action recognition system. The Microsoft Kinect sensor generates three types of data: RGB images, depth maps that gather the distances of different scene objects from the viewpoint, and Skeleton data that is represented by a set of 3D positions of different human body key joints.\\ 
Several approaches were proposed based on each modality: depth approach by extracting features from depth maps and exploring different points \cite{11,12}, skeleton approach that involves the skeleton representation of human body movement where the positions and orientations of the joints are used to describe the human poses and actions \cite{14,15}, and hybrid approach that extract features by combining the two types of data \cite{13}.\\
In this work, we focus on skeleton based approach due to its several advantages over the other modalities such as robustness to view-point and illumination changes, and its compact representation resulting in a reduced computation cost.\\
The initial deep learning skeleton-based HAR systems employ traditional deep learning models like Convolution Neural Networks(CNN) and Recurrent Neural Networks(RNN). To make use of CNNs, researchers convert sequences of skeleton data into pseudo-images in order to have a Euclidean representation, which is then fed to the CNN \cite{16,17}. RNN-based methods, on the other hand, also need a transformation by representing each joint by a sequence of coordinate vectors \cite{18,19}. In fact, skeleton data are embedded as graph-structured data where the joints are the nodes and the bones are the edges linking different joints. Recently, Deep learning methods have been generalized to treat graph-based problems using Graph Neural Network (GNN) to capture both local and global structural information \cite{6}. Several variants of GNN were proposed with different architectures that offer a range of ways to handle information propagation and aggregation on the graph. One popular GNN is the Graph Convolution Network (GCN)\cite{7} which uses a spectral convolution operator to aggregate information from a node's neighbors in the graph. GCNs have been shown to be effective for skeleton-based human action recognition. In \cite{8}, Yan et al. proposed a novel model for skeleton based action recognition: Spatio-Temporal GCN (ST-GCN). It can capture spatial patterns from joints distributions as well as their temporal dynamics. Following the same concept, numerous ST-GCN variants were emerged within the past few years \cite{3,20,35} that achieved significant performance on different benchmarks.

\subsection{Skeleton-based Human Action Recognition with Spatio-Temporal Graph Convolution Networks}

\begin{figure}[]
    \centering
    \includegraphics[scale=0.8]{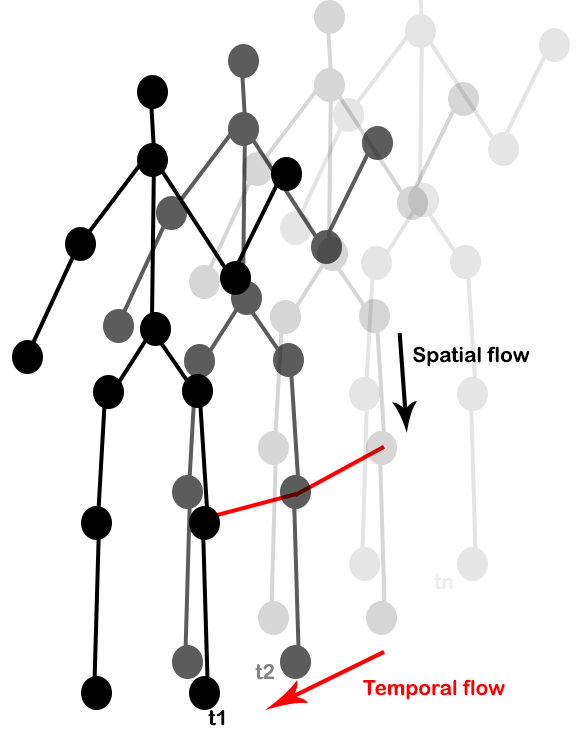}
    \caption{Spatio-temporal skeleton representation:  edges in black are spatial edges and red links are the temporal edges.}
    \label{st_skeleton}
\end{figure}

GCN-based methods showed better ability to capture actions' patterns than CNN-based and RNN-based methods because of the non-Euclidian nature of skeleton representation. Also, it represents a simpler method by eliminating the pre-step of manual data transformation as needed for CNN and RNN. Recently, variant ST-GCNs \cite{3, 20, 35} were proposed following the framework introduced in \cite{8} that can learn both spatial and temporal patterns from the skeleton sequences by extracting features from spatial edges that express connectivity between human joints, and temporal edges that connect the same joints across time steps (Figure.\ref{st_skeleton}).

\subsubsection{Graph-based representation of action skeleton sequence}
ST-GCN models take, as input, a spatio-temporal representation of the skeleton action sequence defined by a set of undirected graphs \{$G_1$,…,$G_n$\} where $G_i$ is the skeleton graph having spatial edges of each frame, and $n$ is the number of frames of the input sequence. The skeleton graph is represented as $G=\{V,E\}$ where $V=\{v_1,…,v_m\}$ is the set of nodes referring to the human body joints. $m$ is the total number of nodes where each has a set of features $X$. $E$ is the set of edges connecting different joints including spatial edges that are bones representing natural connections between the joints in the human body, and temporal edges connecting the joints of two adjacent frames. The graph is denoted by an adjacency matrix $A \in \{0,1\}$ of $m\times m$ dimension, and $A_{i,j}=1$ if $v_i$ and $v_j$ are adjacent and $A_{i,j}=0$ otherwise. 

\subsubsection{Graph Convolution Network} 
The goal of GCN is to learn a new set of features of a given input graph with features $X$, by capturing information from both the node's own features and the features of its neighbors. GCN falls under the category of message-passing neural networks. It can be designed by stacking multiple layers on top of each other, where the output of one layer serves as the input of the next one. The layer-wise update rule applied to features $X$ at time $t$ can be defined as follows: 
\[ X_t^{l+1}= \sigma (\tilde{D}^{\frac{1}{2}}  (A+I) \tilde{D}^{\frac{1}{2}}   X_t^l   W^l  ) \]
$(A+I)$ represents the addition of self-loops to add the current node's features. $\tilde{D}$ is the diagonal degree matrix of  $(A+I)$, $W$ is the network weights and  $\sigma$ is the activation function.

\subsubsection{ST-GCN}
\begin{figure}[h]
    \centering
    \includegraphics[scale=0.8]{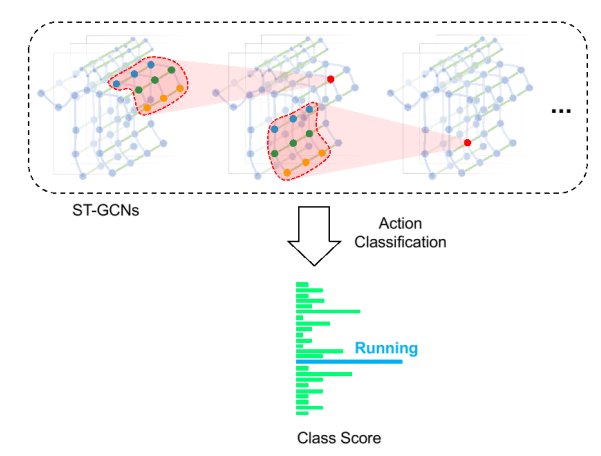}
    \caption{Spatio-temporal Graph Convolution Network architecture \cite{8}: multiple Spatio-temporal Convolution layers stacked to extract features followed by a SoftMax classifier to predict the action label.}
    \label{st_gcn}
\end{figure}

The concept of GCN in the skeleton-based human action recognition was first introduced by Yan et al. \cite{8}. They presented the graph convolution method on spatio-temporal representation of skeleton sequences. This model uses multiple layers (ST-GCN blocks), that calculate spatial and temporal convolutions simultaneously (Figure.\ref{st_gcn}). Mathematically, ST-GCN can be defined as follows:
\[ X^{l+1}= \sum \breve{A} X^l W^l \]
$\breve{A}$ refers to the normalized form of the adjacency matrix $A$: $ \breve{A}= D^{-\frac{1}{2}} AD^{-\frac{1}{2}}$ . Then, a global pooling and softMax function are used to recognize the action of the resulting tensor.

\subsubsection{Two-Stream Adaptive Graph Convolution Network (2s-AGCN)}
\begin{figure}[h]
    \centering
    \includegraphics[scale=0.85]{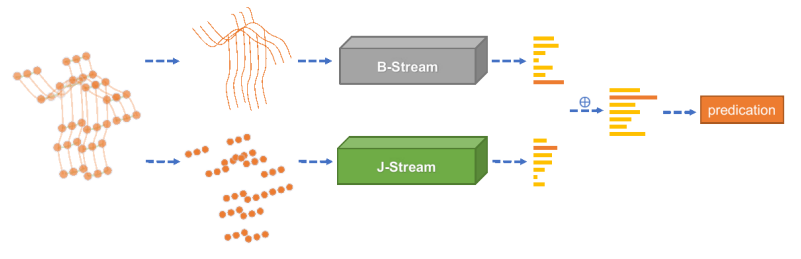}
    \caption{The two-stream AGCN design\cite{3}: B-stream and J-stream are the Adaptive Graph Convolution Neural Networks for Joints and bones respectively.}
    \label{2s_AGCN}
\end{figure}

Shi et al. \cite{3} introduced a new variant of ST-GCN: Two-stream adaptive graph convolution networks. In contrast to the first ST-GCN where the topology of the graph is fixed and set manually, 2s-AGCN learns the human graph topology adaptively during the training process to increase the flexibility of the model in representing skeleton sequences. Moreover, a two-stream framework is used to capture both first-order information: the joint dependencies, and second-order information: the directions and the length of bones (Figure.\ref{2s_AGCN}), in order to boost the performance. It showed a notable improvement in the recognition accuracy on benchmark datasets, and it surpasses the first proposed ST-GCN. The layer-wise update rule of the 2s-AGCN can be defined as:
\[ X^{l+1} = \sum X^l W^l (\breve{A} +\breve{B}+\breve{C}) \]
$A$ is the normal adjacency matrix, $B$ is the learnt adjacency matrix during the training and $C$ is the node similarity matrix. $\breve{A}$, $\breve{B}$, and $\breve{C}$ are the normalized form of $A$, $B$ and $C$ respectively.

\subsubsection{MS-G3D}
Multi-scale spatio-temporal GCN is another variant of GCNs for skeleton based action recognition task \cite{4,5}. It is based on a feature aggregation from long-range neighbor nodes using higher polynomials of the graph adjacency matrix. However, this aggregation with adjacency powering can make it ineffective to capture long-range joint dependencies because of the weighting bias problem where the aggregated feature will be dominated by signals from local body parts. Liu et al. \cite{35} proposed a disentangled and unifying GCN framework to encounter this problem. They introduced a disentangled multi-scale aggregator that obtains direct information from farther nodes and removes redundant dependencies between node features, using a k-adjacency matrix as:
\[
    [\tilde{A}_k]_{i,j}= 
\begin{dcases}
        1 ,& \text{if } d(v_i,v_j)=k\\
    1 ,& \text{if } i=j\\
    0,              & \text{otherwise}
\end{dcases}
\]
where $\tilde{A}=A+I$, $k$ is the number of scales to aggregate, and  $d(v_i,v_j )$ returns the shortest distance in number of hops between $v_i$ and $v_j$. 
Moreover, this model uses a unified spatial-temporal graph convolution operator to facilitate direct information flow across space and time. The combination of these two methods results in a powerful feature extraction across both spatial and temporal dimensions. It significantly outperforms the state-of-the-art methods. 

\subsection{Transfer learning}
Transfer learning is a deep learning strategy that has been successfully applied to a wide range of applications. It consists of taking past learnt knowledge, in a specific context: \textit{source domain}, and reusing it to solve a new problem in a related context: \textit{target domain}. Usually, deep learning models are trained from a random weights initialization, for a specific task using a large number of training data labels. If the task changes, the model must be retrained from scratch. In transfer learning, the retraining is applied using a previously-trained network’s weights, rather than retraining from scratch. This technique can be especially useful when there are limited training data labels of the new task, as the pre-trained model can help to enhance the performance with fewer samples. Additionally, transfer learning can help to speed up the training process and reduce the computational resources needed for training a new model from scratch. There are two ways of applying transfer learning: 
\begin{itemize}
    \item Fixed-weights transfer: using the pre-trained model's weights as fixed features, where all the weights of different layers, except the last one, are frozen. 
    \item Fine-tuned transfer: using the pre-trained model's weights as the starting point, where all network's weights can change during the retraining.
\end{itemize}
With the recent explosion in the number of HAR applications and the success achieved by transfer learning technique, many researchers have been investigating transfer learning for HAR task \cite{21,22,23}. However, most of the proposed works were based on RGB images using Euclidean Neural Networks. For instance, in \cite{21}, CNN and LSTM (Long Short Term Memory) were used to develop an HAR framework based on the transfer learning technique using RGB images.\\
To our knowledge, no works have been interested in exploring transfer learning for 3D skeleton-based HAR. Additionally, there has been little insight into the transferability of GNNs and not much research exit that investigates transfer learning for GNNs. Although, in \cite{24}, the effectiveness of transfer learning with GNNs has been demonstrated and the experiments on real-world datasets showed that the transfer is most effective when the source and target graphs are similar.\\

The main contributions of this work are:
\begin{itemize}
    \item We present a novel dataset comprising of skeleton sequences of human actions recorded within a theatre scene environment, utilizing the Kinect sensor. This dataset serves as the foundation for developing the first-ever theatre Human Action Recognition (HAR) framework.
    \item We thoroughly investigate and analyze the transferability of three widely-used variants of Spatio-Temporal Graph Convolution Network (ST-GCN) models for the skeleton-based HAR task. Specifically, we examine ST-GCN \cite{8}, 2s-AGCN \cite{3}, and MS-G3D \cite{35}. Through our experiments, we observe positive transfer effects as all the models demonstrate improved performance compared to the baseline approach without transfer learning.
    \item In addition, we introduce a framework that enhances the performance of transfer learning methods by effectively addressing the differences between the target domain (our dataset) and the source domain (NTU-RGBD dataset\cite{1,2}: a benchmark for human action recognition). This framework leads to significant performance improvements in the HAR task.
\end{itemize}

\section{Problem Positioning}
\begin{figure}[ht]
    \centering
    \includegraphics[scale=1.2]{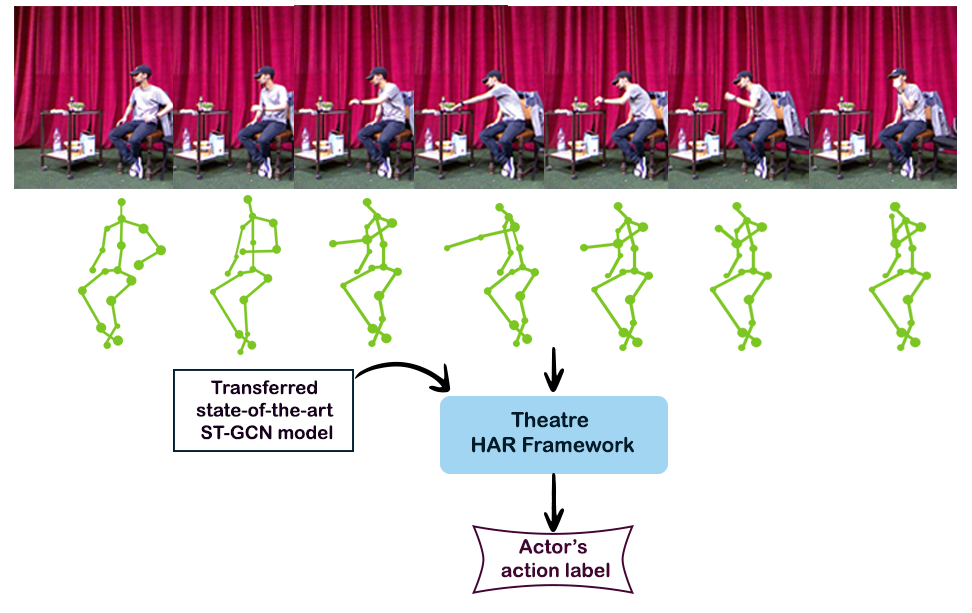}
    \caption{(Top) The Actor's action on stage, (middle) The corresponding skeleton sequence provided by Kinect v1 (Bottom) The components of the recognition system: It takes as input the skeleton sequence and exploits ST-GCN models with transfer learning to recognize the performed action.}
    \label{schema}
\end{figure}

The goal of this research work, as mentioned before, is to develop a HAR system that recognizes the actions of actors in a theatre scene in order to contribute to the development of an output device that helps Visually Impaired and Blind persons understand theatre scenes. The device takes as input the actor's action label recognized by a theatre HAR system and provides the corresponding action description to the user as output.\\ 
In this work, we focus on elaborating the theatre HAR system. To accomplish this, we rely on the skeleton representation of the performed actions provided by Kinect sensor, and we exploit the state-of-the-art ST-GCN models using the transfer learning technique (as shown in Figure.\ref{schema}).
\\
We choose to employ the skeleton approach with ST-GCN models using transfer learning due to the following reasons: 
\begin{itemize}
    \item The first choice is justified by the fact that Skeleton approach has achieved considerable success over the other approaches due to its several advantages: 
        \begin{itemize}
        \item It is robust to occlusion: It uses only the joint positions, which can be estimated even when some body parts are occluded.
        \item It is invariant to appearance changes: The skeleton-based approach is invariant to changes in clothing, illumination conditions, and camera viewpoint, due to the use of joint positions only which are relatively stable under these changes.
        \item It is computationally efficient: The compact skeleton representation of the human body helps reduce the computation time which is crucial in the case of real-time applications.
        \end{itemize}
In addition, the skeleton representation provides better both spatial and temporal information with the strong intra-frame and inter-frame correlations between the joints allowing the recognition algorithms to capture more expressive and significant features.\\
The skeleton data can be acquired using different techniques, each has its advantages and disadvantages. In   Table.\ref{skeleton_tech}, we list these techniques and provide a comparison between them after analysing their characteristics in the context of theatre.

\begin{table}
\begin{center}
\begin{tabularx}{1\textwidth}
{ | >{\raggedright\arraybackslash}X 
  | >{\raggedright\arraybackslash}X 
  | >{\raggedright\arraybackslash}X |} 
  \hline
  Technique& Advantages & Disadvantages \\ 
  \hline
  \hline
  & & \\
  Motion capture technologies using sensing devices. & 
  It provides very precise annotations of skeleton data. &
  It is hard to implement because the actors will have to wear sensing devices which are expensive and can hamper the actors' performance on the stage. \\ 
  & & \\
  \hline
  & & \\
  Collecting the skeleton data using Pose Estimation method\cite{28} with RGB images. & 
   It is simple to implement using only an RGB camera.  
    Not expensive. & 
    Its precision can be affected by several factors such as illumination variation, clothing colors, and complex backgrounds.\\
    &&It is impossible to control these factors in a theatre environment.\\
    &&It does not provide the depth information which can significantly boost the performance of the HAR system.\\ 
    & & \\
  \hline
  & & \\
  Using Microsot Kinect sensor. &
  It captures both RGB and depth information. This combination allows it to produce very accurate joints estimation. &  It is limited by the distance: It cannot capture depth information from a large distance.\\
 & Not expensive. & \\
 & & \\
\hline
\end{tabularx}
\caption{Comparison between the skeleton data generation techniques in the context of theatre scene.}
\label{skeleton_tech}
\end{center}
\end{table}

According to Table.\ref{skeleton_tech}, the Kinect sensor is the most suitable device for the theatre HAR system, which provides rich information and does not affect the actors' performance. The limitation of distance can be encountered by placing the Kinect sensor in a near position to the stage.

    \item The second choice is justified by the fact that the GCN-based approach provides the ability to model and learn joint dependencies across space-time implicitly, unlike RNN-based and CNN-based approaches where a data transformation step is necessary to represent skeletons as 2D or 3D Euclidean grids to be fed into the model (RNN or CNN). In addition, GCN has outperformed CNN and RNN in many recent studies on HAR. Most recent HAR research works follow the framework introduced by Yan et al.\cite{8} using the spatio-temporal modeling of skeleton data.
    \item We use transfer learning despite the success achieved by the ST-GCN models on challenging human action benchmarks, because the use of one of the trained models with new users can decrease its performance if the action patterns of the new user are different from those in the training data. In our case, actions' patterns in theatre scene environment can be different from the patterns of the existing human action benchmarks. A typical solution is to collect a new dataset containing theatre human action samples and train the model from scratch. This task is not feasible due to the following reasons:
        \begin{itemize}
        \item Deep learning models require large amount of training samples to learn intricate patterns and generalize well to new domains, which makes the task of manually collecting a new dataset very hard and time-consuming especially in the case of theatre actions. It is not convenient to get access to theatres and collect a large number of samples that can take several days to accomplish.
        \item The training of a deep learning model is time-consuming and requires high computation energy which in turn contributes to carbon emissions. This energy demand has seen immense growth in recent years and deep learning may become a significant contributor to climate change if this trend continues. As a result, the excitement over Deep learning success has shifted to warning and many recent studies showed interest in the environmental impact of deep learning, encouraging research into energy-efficient approaches by taking simple steps to reduce carbon emissions \cite{25,26}. 
        \end{itemize}
To encounter the cited problems, we investigate transfer learning technique on ST-GCNs for skeleton-based HAR. It helps tackling the problem of limited training data labels and it can also significantly reduce training time leading to decrease the energy and the carbon emissions. Thus, we adopt the transfer learning technique, not only because of the training data scarcity problem, but also to promote responsible computing and to avoid the cost of training models for extensive periods on specialized hardware accelerators.\\
In addition, while conventional deep learning models, such as CNNs, have demonstrated remarkable transferability, research analyzing transfer learning technique within the graph-based field and their ability to transfer learnt knowledge with GNNs is limited. Furthermore, to our knowledge there are currently no studies examining transfer learning via spatio-temporal GCNs for skeleton-based human action recognition.
\end{itemize}

\section{The Proposed Method}
\subsection{Our Theatre Human Actions Dataset}
\begin{figure}
    \centering
    \includegraphics[scale=0.03]{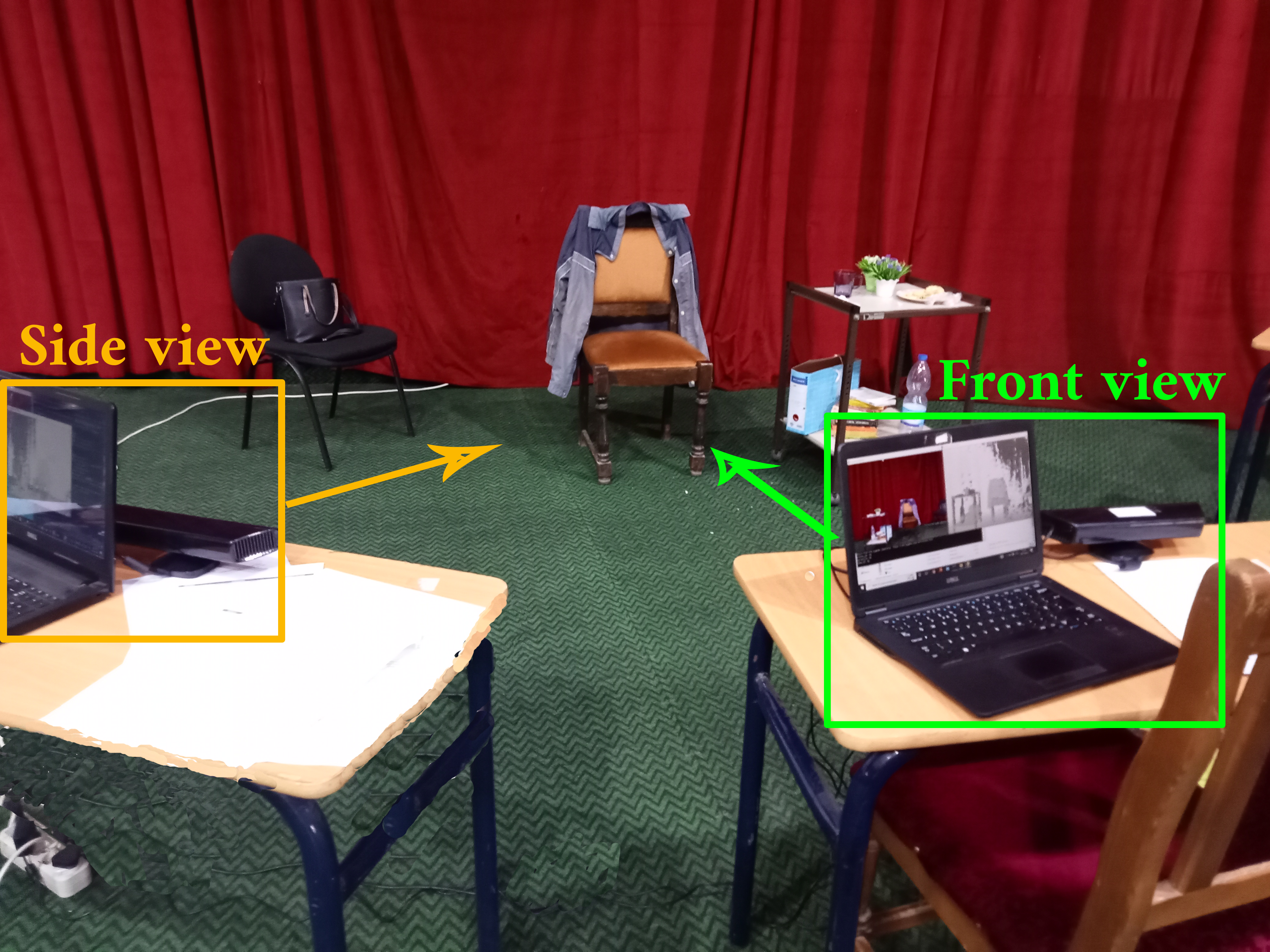}
    \caption{The setup of the scene and the placement of the two cameras: one camera from the front view and a second camera from the side view}
    \label{setup}
\end{figure}

\begin{figure}
    \centering
    \includegraphics[scale=0.2]{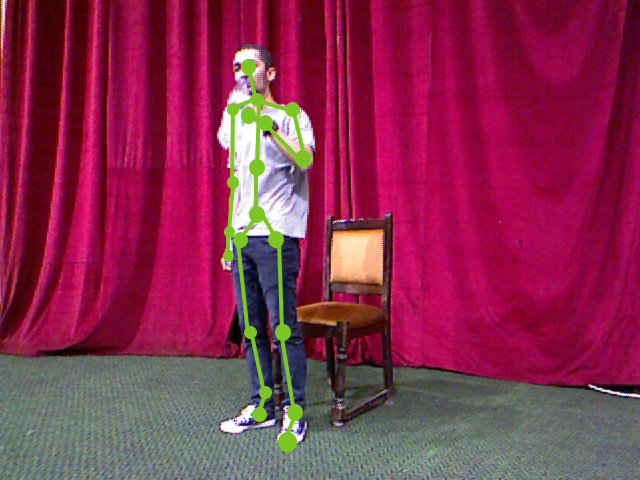}
    \includegraphics[scale=0.2]{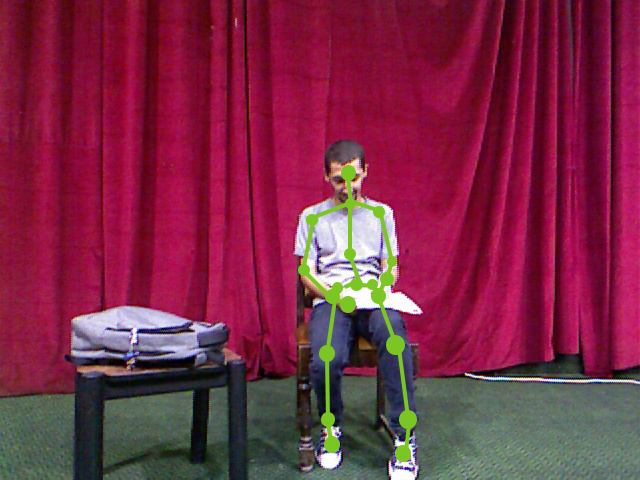}
    \includegraphics[scale=0.2]{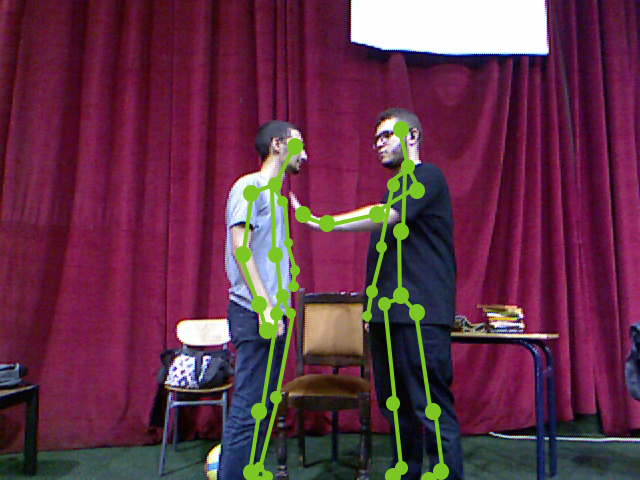}
    \caption{ Samples from our dataset:(Top-left) Drinking. (Top-right) Writing,
    (Bottom) Pushing a person.}
    \label{samples}
\end{figure}

\begin{figure}
    \centering
    \includegraphics[scale=0.8]{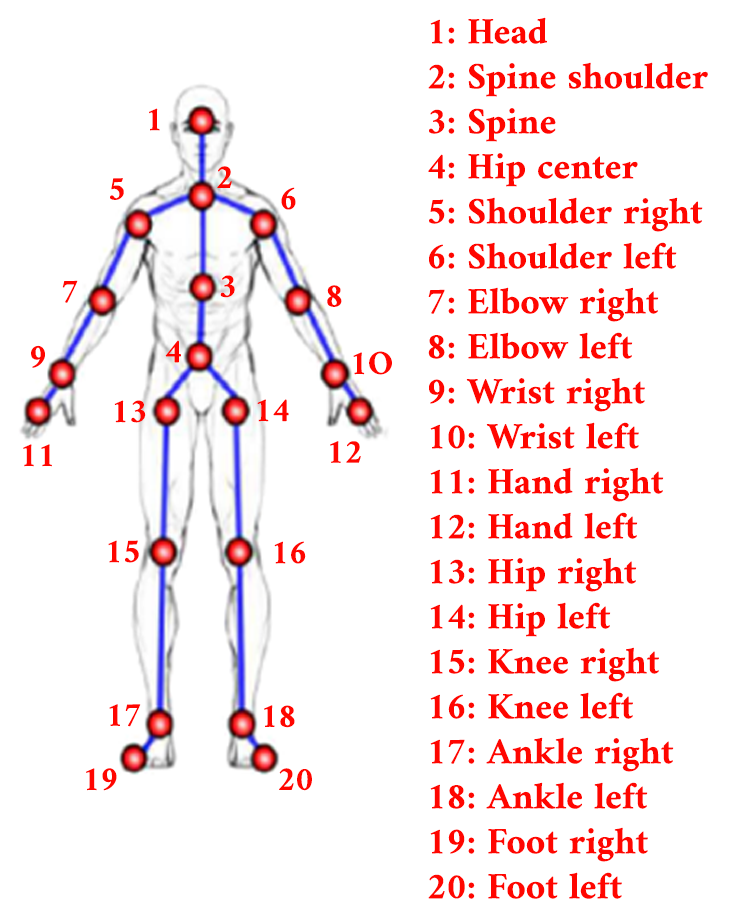}
    \caption{Configuration of skeleton joints captured by Kinect v1}
    \label{v1_config}
\end{figure}

We collected a new dataset that contains skeleton action sequences at the auditorium of our university where two Kinect v1 are positioned at the same height and in two different view angles (front view and side view) resulting in a variety of the obtained data (Figure.\ref{setup}).
\\
Each action was performed by $3$ individuals and repeated at least $3$ times with different movement speeds. In total, we collected $230$ sequences from each viewpoint over $36$ action classes, with a rate of $25$ frames per second and an average of $170$ frames per sequence. The collected action classes are actions that are most accurate in theatre scenes including solo actions such as \textit{walking}, \textit{sitting down}, \textit{jumping}, \textit{throwing}, \textit{drinking}, and two-person interactions such as \textit{hugging}, \textit{kicking a person}, \textit{shaking hands}, \textit{giving an object to a person}. Some samples from our dataset are illustrated in Figure.\ref{samples}.
\\
The provided skeleton information consists of $3$D positions of $20$ body key joints for each tracked human body. The configuration of the $20$ joints is illustrated in Figure.\ref{v1_config}.

\subsection{Transfer Learning}
\begin{figure}[h]
    \centering
    \includegraphics[scale=1.2]{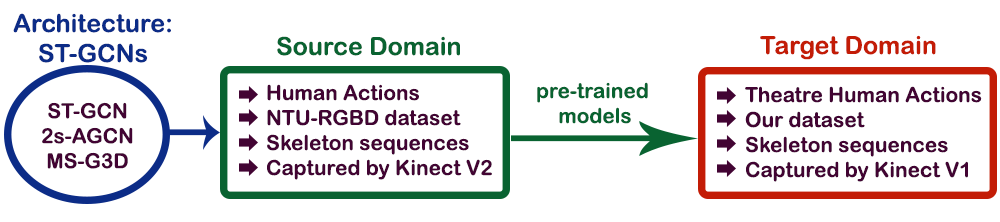}
    \caption{Our transfer learning investigation parameters: The chosen architectures of the ST-GCN models (blue circle) : ST-GCN, 2S-AGCN, and MS-G3D. They are trained on a source domain (green rectangle): Human action skeleton sequences of NTU-RGBD captured by Kinect v2. Then, they are transferred to a target domain (red rectangle):  Theatre human skeleton sequences of our dataset that are captured by Kinect v1.}
    \label{tl_schema}
\end{figure}

Limited research has been conducted on the transferability of GCNs despite the growing interest in them. The present study aims to explore the transferability of spatio-temporal GCNs for recognizing human action based on skeletons. However, the performance achieved by transfer learning relies on the choice of the pre-trained model and the source domain. This task remains difficult due to the diversity of state-of-the-art architectures. Thus, we selected three of ST-GCN common models (defined in section \textbf{\textit{2}}): the first proposed ST-GCN \cite{8}, the two-stream adaptive GCN \cite{3}, and the multi-scale disentangled unified GCN \cite{35}, because they showed a significant performance and are open source and their pre-trained models are available. Furthermore, according to \cite{9}, the choice of the source domain can be made based on its size and its similarity with the target domain. Since we are focusing on a skeleton-based approach, the NTU-RGBD dataset represents a leading option due to its popularity and data diversity. In addition, its structure is quite similar to our collected dataset since both were captured using Kinect sensors and both contain indoor actions.\\ 
Figure.\ref{tl_schema} illustrates different inputs and parameters of our transfer learning investigation.

\subsubsection{Source Domain: NTU-RGBD Dataset}
\begin{figure}[h]
    \centering
    \includegraphics[scale=0.85]{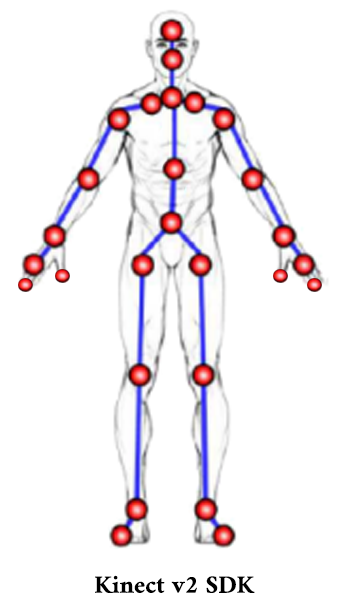}
    \caption{Skeleton joints provided by Kinect v2}
    \label{kinectv2}
\end{figure}

It represents one of the most challenging and largest benchmarks due to the diversity of its data. It was captured using Kinect v2 which provides RGB, Depth, and skeleton sequences. The provided skeleton data consists of $25$ human body joints (Figure.\ref{kinectv2}). NTU60-RGBD \cite{1} was first introduced containing $60$ indoor human action classes and a total of $56880$ samples performed by $40$ subjects with $80$ different camera setups. Then, the extended version NTU120-RGBD \cite{2} was introduced which contains additional $57.367$ sequences of $60$ extra indoor action classes with a total of $113.945$ samples over $120$ classes captured from $32$ different camera setups and $106$ subjects. The action classes are divided into three categories:  daily actions (e.g.\ walking, reading, phone call …etc), medical conditions (e.g.\ sneezing/coughing, staggering, falling down …etc), and two-person interactions (e.g.\ pushing, punching, hugging …etc).
\\
The authors of this dataset proposed two protocols for recognition evaluation: xview protocol and xsub protocol:
\begin{description}
    \item [The xview protocol] it focuses on cross-view action recognition. Its goal is to train a model on one set of viewpoints and evaluate its performance on a different set of viewpoints. Hence, the dataset is divided into two parts: \textit{i) training set} that contains samples captured from one set of camera viewpoints, and \textit{ii) testing set} that contains samples from a different set of camera viewpoints.
    
    \item [The xsub protocol] it focuses on cross-subject recognition where the objective is to train a model on a specific set of subjects and evaluate its performance on unseen subjects. The training set includes sequences of certain subjects, while the testing set consists of sequences of different subjects.
\end{description}
The three models: ST-GCN, 2s-AGCN, and MS-G3D achieved good results on this benchmark as shown in Table.\ref{ntu_perf}. Their high obtained performances on both protocols (xsub and xview) demonstrate their ability to achieve robust and generalized human action recognition across different subjects and camera viewpoints.
\begin{table}  
\begin{center}
\begin{tabularx}{0.7\textwidth}
{ | >{\raggedright\arraybackslash}X 
  | >{\centering\arraybackslash}X 
  | >{\centering\arraybackslash}X |} 
    \hline
   & \multicolumn{2}{|c|}{NTU60-RGBD } \\
 \hline
 & xsub & xview \\
 \hline
 ST-GCN \cite{8} & $81.5\%$ & $88.3\%$  \\
 \hline
 2s-AGCN \cite{3} & $88.5\%$ & $95.1\%$ \\
 \hline
 MS-G3D \cite{35} & \textbf{91.5\%} & \textbf{96.6\%}\\
 \hline
\end{tabularx}
\caption{The obtained accuracies by the selected ST-GCN models on NTU60-RGBD following the two evaluation protocols: xsub and xview.}
\label{ntu_perf}
\end{center}
\end{table}

\subsubsection{Diversity and similarity between source and target domains}
\begin{figure}
    \centering
    \includegraphics[scale=1.1]{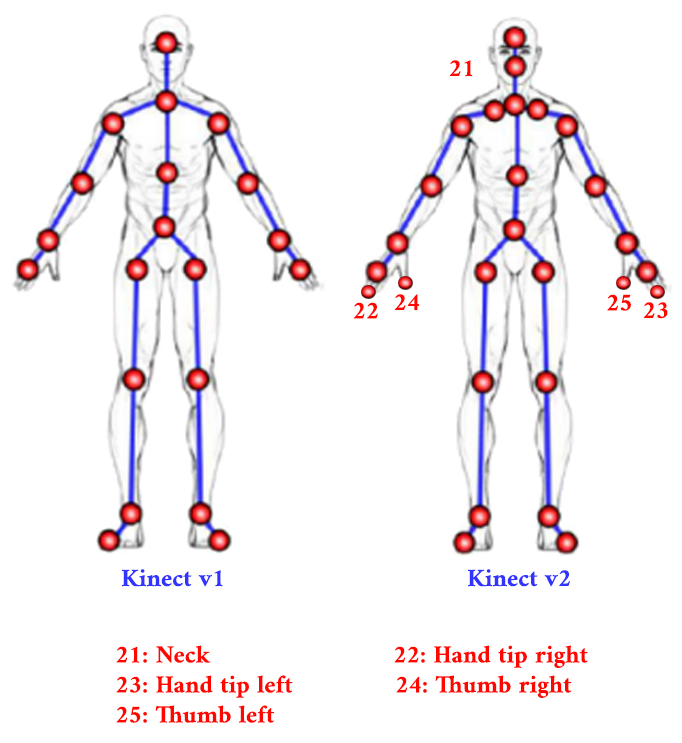}
    \caption{The five additional joints captured by Kinect v2}
    \label{add_joints}
\end{figure}

\begin{figure}
    \centering
    \includegraphics[scale=0.8]{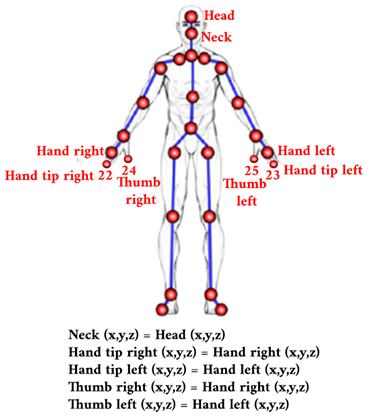}
    \caption{The transformation of 20 joints skeleton to 25 joints skeleton.}
    \label{duplication}
\end{figure}

Our dataset, as mentioned above, was captured using Kinect v1, which provides 3D positions of $20$ body joints. Unlike the NTU-RGBD dataset, it was captured using Kinect v2 which provides 3D positions of $25$ joints. The additional five joints are: \textit{neck}, \textit{hand tip right}, \textit{hand tip left}, \textit{thumb right} and \textit{thumb left}, as shown in Figure.\ref{add_joints}.
Since these joints are not distant from the joints: \textit{head}, \textit{hand right}, \textit{hand left}, \textit{hand right} and \textit{hand left} respectively, and the links between them do not represent body bones, we duplicated the positions of their neighboring joints in order to transform our dataset to $25$ joints dataset and add the positions of the five missing joints (as shown in Figure \ref{duplication}). This transformation allows us to save the same graph structure used in the training of the pre-trained models. 
\\
In addition, after we examined both datasets, it was evident that the actions in our dataset were executed at a slower pace when compared to the NTU action sequences. The average number of frames per sequence in our dataset is around $170$ frames, while the average in NTU sequences is around $95$ frames.  Hence, we propose a framework for transfer learning application in section \textit{\textbf{5.2.4}} to adapt the models to the new temporal correlations.

\section{Validation and Discussion}
\subsection{Training}
We first trained the selected models on our dataset in order to calculate the performance of transfer learning and be able to compare it with the baseline. We utilized the identical architectures and parameters used while their training on NTU-RGBD: 
\begin{itemize}
    \item \textbf{ST-GCN} is comprised of $9$ layers of ST-GCN units (spatial temporal graph convolution operators) followed by a global pooling to get a feature vector for each sequence and feed it to a SoftMax classifier \cite{8}. We use the stochastic gradient descent: Adam optimizer, for the learning process of the model with a learning rate of $0.1$ which is decayed by $0.1$ after every $20$ epochs.
    \item \textbf{2s-AGCN} is composed of two adaptive graph convolution networks: J-stream and B-stream representing the networks of joints and bones, respectively (as shown in section \textit{\textbf{2.1.1}}).  Each network has a total of $9$ blocks followed by a global average pooling layer to pool feature maps of different sequences to the same size and feed it to a SoftMax layer. Then, the scores of the SoftMax classifiers of both J-stream and B-stream are fused to predict the action label \cite{3}.  The learning rate is fixed at $0.1$ and decayed by $0.1$ every $20$ epochs starting from epoch number $30$.
    \item \textbf{MS-G3D} contains a stack of $3$ spatial-temporal graph convolutional (STGC) blocks followed by a global average pooling layer and a SoftMax classifier. Each STGC block is composed of two types of pathways to simultaneously capture long-range spatial and temporal dependencies using multi-scale convolutional layers, as well as regional spatial-temporal joint correlations by performing disentangled multi-scale convolutions. Then, the outputs from all pathways are aggregated as the STGC block output \cite{35}. The learning rate is initiated at $0.5$ and it is divided by $10$ after every $30$ epochs starting from epoch number $10$.  
\end{itemize}
Since the max number of frames in each action sequence in our dataset is $300$ frames, all the samples with less than 300 frames are padded by replaying the sequences until they reach $300$ frames. Then, we apply normalization and translation on the samples following \cite{29,3}.\\
All the experiments in this research were repeated numerous times in order to ensure the validity of our findings, and the performance averages are reported.\\
The obtained results, illustrated in Table.\ref{tha_perf}, show a low performance on our dataset which was expected due to: \textbf{1)} the training data scarcity problem that prevents the models from capturing sufficient spatial and temporal patterns from the new dataset, \textbf{2)} the low precision of the provided joint positions by Kinect v1 compared to Kinect v2.
\begin{table}  
\begin{center}
\begin{tabularx}{0.65\textwidth}
{ | >{\raggedright\arraybackslash}X 
  | >{\centering\arraybackslash}X 
  |} 
    \hline
   & Accuracy \\
 \hline
  
 ST-GCN & $34.48\%$  \\
 \hline
 2s-AGCN & $39.66\%$  \\
 \hline
 MS-G3D & $41.38\%$ \\
 \hline
\end{tabularx}
\caption{The obtained accuracies by the ST-GCN models on our dataset.}
\label{tha_perf}
\end{center}
\end{table}

\subsection{Transfer Learning}
In this section, we present our study of the effectiveness of transfer learning technique with the selected ST-GCNs using two frameworks: \textit{configuration1} and \textit{configuration2}.\\
\textit{Configuration1} represents the implementation of the fixed-weights transfer defined in section \textit{\textbf{2.2}}. Next, in order to adapt the pre-trained models to the target domain's temporal correlations, we propose a framework: \textit{configuration2}, that combines the two transfer learning approaches: fixed-weights and fine-tuned transfer. The results are reported and discussed.

\subsubsection{Evaluation Metrics}
\begin{figure}[h]
    \centering
    \includegraphics[scale=0.8]{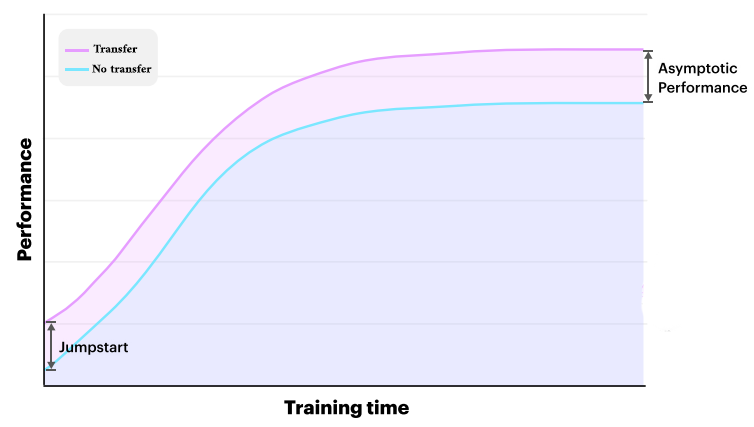}
    \caption{Illustration of the jumpstart and the asymptotic performance\cite{24}: The blue curve represents the baseline without transfer learning and the purple curve refers to the transferred model from a source domain.}
    \label{eval}
\end{figure}

In order to evaluate the performance of the transfer learning technique, the jumpstart and asymptotic metrics, introduced by Taylor and Stone\cite{10}, were employed to assess the transferability of the pre-trained models performance. \textbf{Jumpstart} is defined as the difference between a model's initial performance in a target task and its initial performance after being transferred from a source task. \textbf{Asymptotic performance}, on the other hand, measures the improvement in the final performance achieved in the target task through transfer learning, compared to the performance achieved by the baseline (as shown in Figure.\ref{eval}).
\subsubsection{Configuration1}
\begin{figure}[ht]
    \centering
    \includegraphics[scale=1.3]{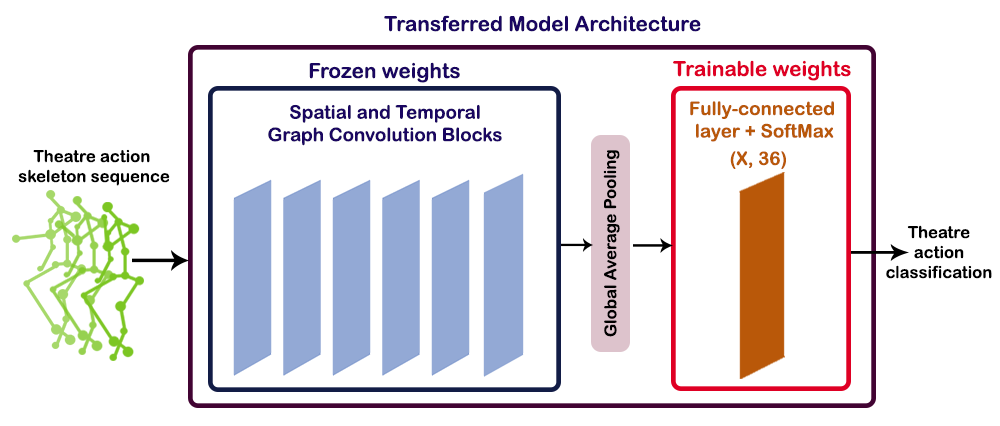}
    \caption{An illustration of the framework of configuration1 on the models showing the trainable and the frozen blocks, as well as the added fully-connected layer with X representing the output feature tensor and 36 referring to the number of labels in the target domain.}
    \label{conf1}
\end{figure}

As a first attempt, we implemented the fixed-weights transfer approach due to the similarity between the graph structure of both source and target domains after applying the data pre-processing method detailed in section \textbf{\textit{4.2.2}}. This graph-structure similarity results in a similarity of the spatial patterns that can be captured by the models from both datasets. Therefore, we saved the same architectures and parameters of the pre-trained models listed in section \textbf{\textit{5.1}}. We loaded the weights of the pre-trained models while keeping the same input layer and setting all spatio-temporal convolution blocks to non-trainable in order to preserve the weights learned from the source domain. Then, we modified the output layer of the models by adding a new fully-connected layer to match the $36$ action classes present in our dataset (as shown in Figure.\ref{conf1}). Since all the models' layers are frozen and we are training only the fully-connected layer, we avoid to update the weights with large learning rates to prevent overfitting. Thus, we reduced the learning rate for each model by ten times with a decay of $0.1$ after every $10$ epochs for a stable training. Finally, we initiated the retraining process on our dataset.
\begin{figure}
\begin{center}
    \includegraphics[scale=0.6]{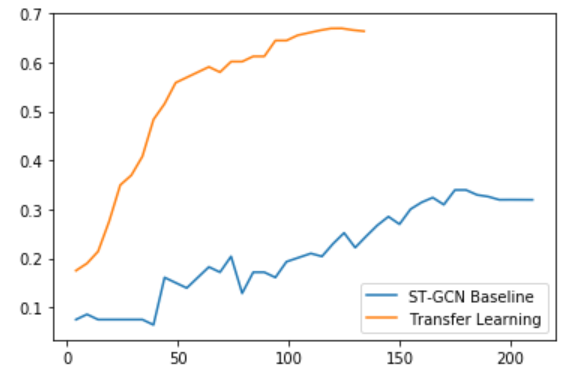}
  \includegraphics[scale=0.6]{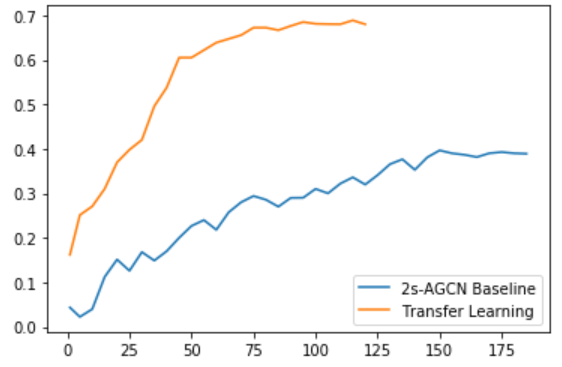}
  \includegraphics[scale=0.6]{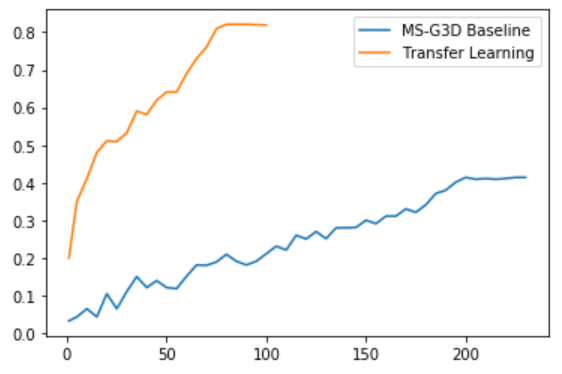}
    \caption{The charts illustrate the accuracy progressions of different transferred models after applying configuration1 and their baselines without transfer learning.}
    \label{charts1}   
\end{center}
\end{figure}
\begin{table}  
\begin{center}
\begin{tabularx}{0.9\textwidth}
{ | >{\raggedright\arraybackslash}X 
  | >{\centering\arraybackslash}X 
  |>{\centering\arraybackslash}X 
  |>{\centering\arraybackslash}X 
  |} 
    \hline
   &Final achieved performance&Jumpstart& Asymptotic performance \\
 \hline
  
 ST-GCN & $0.67$ & $0.10 $ & $0.33 $  \\
 \hline
 2s-AGCN & $0.68$ & $0.12$ & $0.29 $\\
 \hline
 MS-G3D & $0.82$& $0.17$ & $0.41 $\\
 \hline
\end{tabularx}
\caption{The performances (final achieved accuracy, jumpstart, and the asymptotic performance) obtained by the ST-GCN models after implementing transfer learning following configuration1.}
\label{tl_conf1}
\end{center}
\end{table}

\subsubsection{Discussion1}
Table.\ref{tl_conf1} illustrates the calculated metrics and the final achieved performances after applying the transfer learning on each of the three models. From these results and the charts in Figure.\ref{charts1}, we observe that the three models obtained significant jumpstarts and asymptotic performances which prove their ability to transfer past knowledge and demonstrate that spatio-temporal graph neural networks can leverage strong properties in the source task for effective transfer, even with a less precise target dataset. The MS-G3D achieved the best transfer and converged faster compared to the other models, according to the charts  (Figure.\ref{charts1}), which indicates that its architecture allows it to learn powerful transferable spatial and temporal patterns from the source domain resulting in the best transferability. On the other hand, ST-GCN achieved lower performance while training compared to 2s-AGCN as indicated in Table.\ref{tl_conf1} , although, it achieved better asymptotic performance. This shows that ST-GCN benefits better than 2s-AGCN from sharing knowledge from the source domain and that its output classifier could adapt better to the target domain. In addition, the training time has been significantly reduced by the fact that all the models achieved their best accuracies before epoch number $100$ compared to their baselines where they took more than $200$ epochs.  Furthermore, saving time results in the reduction of energy and carbon footprints of the models when training on a new large dataset.

\subsubsection{Configuration2}
\begin{figure}[h]
    \centering
    \includegraphics[scale=1.2]{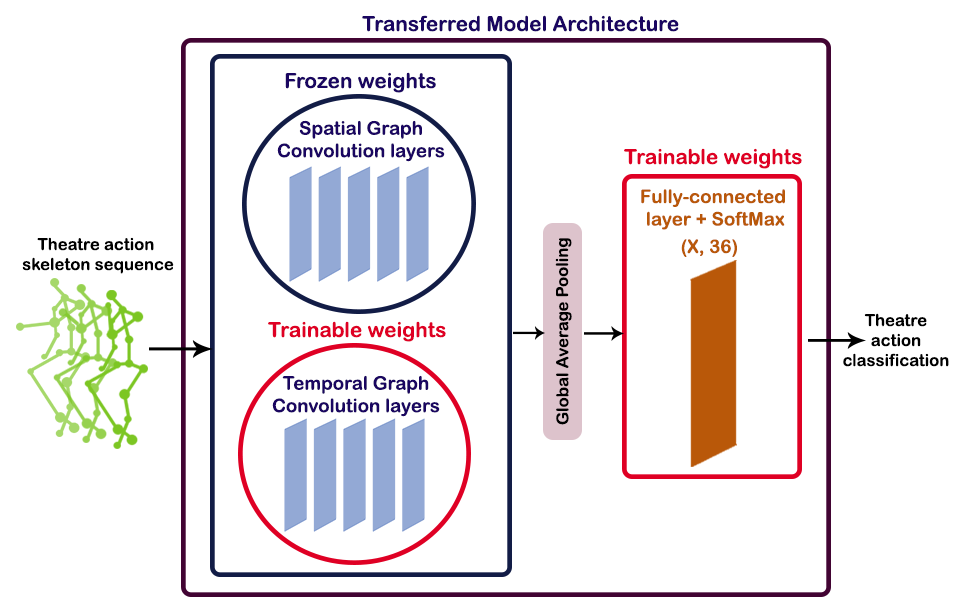}
    \caption{An illustration of the configuration2 on the models showing the trainable blocks: temporal graph convolution layers + fully-connected layer,  and the frozen blocks: spatial convolution layers}
    \label{cong2}
\end{figure}

In order to adapt the transfer learning technique to the temporal diversity between source and target domains introduced in section \textbf{\textit{4.2.2}}, we proceeded with fine-tuning the temporal convolution blocks of each model by enabling them to be trainable so as to enhance the models' ability to learn more temporal features from the target domain. Therefore, we propose a framework that combines the two approaches of transfer learning by applying the fixed-weights approach with the spatial convolution blocks and fine-tuning the temporal convolution blocks (as shown in Figure.\ref{cong2}). We first load the weights of the pre-trained models, then, we freeze the weights of the spatial graph convolution layers of each model. We also change the output layers to match the number of labels in our dataset ($36$ labels), same as in configuration1. We preserve the same parameters of the last experiments except for the learning rate.  A low learning rate in this case is crucial as we are retraining a larger number of weights (temporal convolution blocks) on our dataset which is typically very small. This can lead to an overfitting if we apply large weight updates. To overcome the overfitting, we lower the learning rates by $0.01$ with a decay of $0.1$ every $10$ epochs. Finally, we lunch the retraining of the temporal convolution blocks as well as the output layer.
\begin{figure}
\begin{center}
    \includegraphics[scale=0.6]{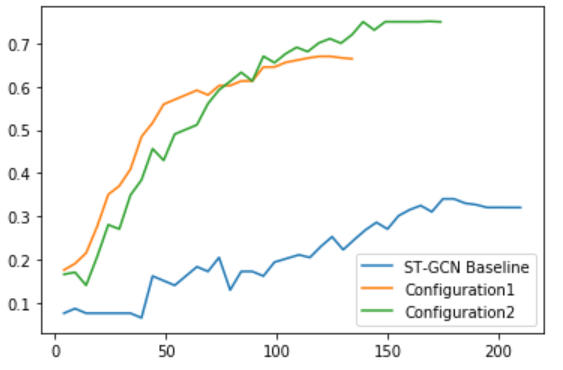}
  \includegraphics[scale=0.6]{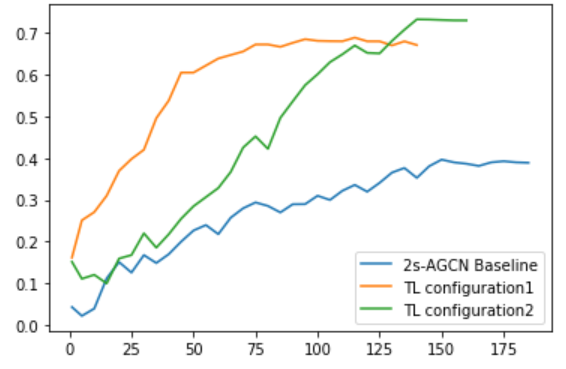}
  \includegraphics[scale=0.6]{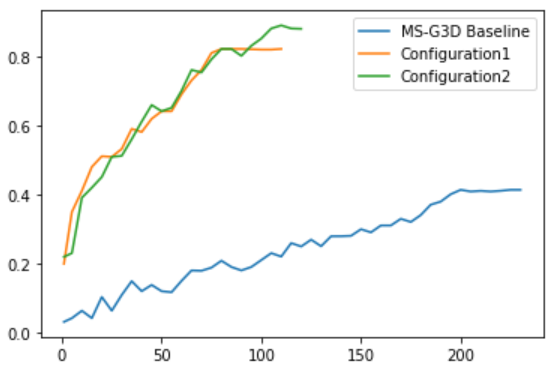}
    \caption{For a clear comparison between the results obtained by configuration1 and configuration2, these charts illustrate the accuracy progressions of different transferred models of both configuration as well as their baselines.}
    \label{charts2}   
\end{center}
\end{figure}
\begin{table}  
\begin{center}
\begin{tabularx}{0.9\textwidth}
{ | >{\raggedright\arraybackslash}X 
  | >{\centering\arraybackslash}X 
  |>{\centering\arraybackslash}X 
  |>{\centering\arraybackslash}X 
  |} 
    \hline
   &Final achieved performance& Jumpstart& Asymptotic performance \\
 \hline
  
 ST-GCN & $0.75$ & $0.09 $ & $0.41 $  \\
 \hline
 2s-AGCN & $0.73$ & $0.11$ & $0.35 $\\
 \hline
 MS-G3D & $0.88$& $0.19$ & $0.47 $\\
 \hline
\end{tabularx}
\caption{The performances (final achieved accuracy, jumpstart, and the asymptotic performance) obtained by the ST-GCN models after implementing transfer learning following configuration2.}
\label{tl_conf2}
\end{center}
\end{table}
\begin{figure}
\begin{center}
    \includegraphics[scale=0.4]{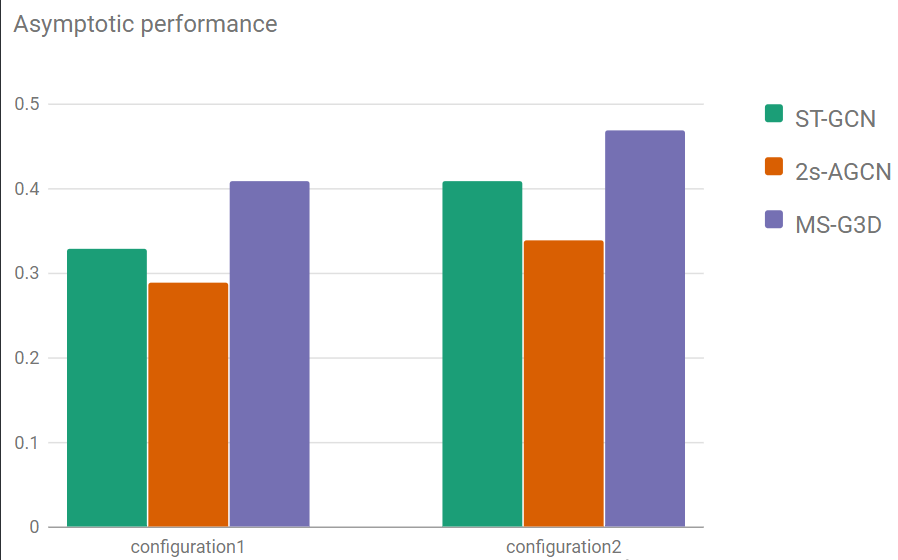}
  \includegraphics[scale=0.45]{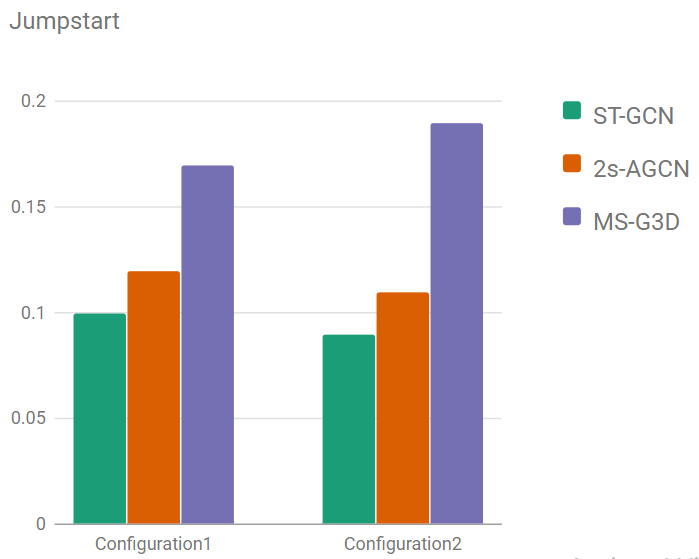}
    \caption{Bar charts to compare the achieved jumpstart values and the asymptotic performances after applying configuration1 and configuration2.}
    \label{chartBars}   
\end{center}
\end{figure}
\subsubsection{Discussion2}
From Table.\ref{tl_conf2}, the charts in Figure.\ref{charts2}, and the bar charts in Figure.\ref{chartBars}, we notice an improvement in the asymptotic performances obtained by all three models while the jumpstart values remain quite similar compared to the previous configuration. This indicates that the models were capable of acquiring new temporal properties from the target domain. It also demonstrates the vital role of selecting the right transfer learning approach for enhanced performance. The understanding of the similarity and diversity among the source and target domains is a key clue to determining which is the most suitable transfer learning approach to implement for better recognition performance. Although, we distinguish that the learning process of this second configuration takes more time to converge. As we can see from the charts in Figure.\ref{charts2}, during the first epochs, the green curves (configuration2) of the three models are under orange curves (configuration1). This is due to the training process of temporal graph convolution layers as they are learning new temporal patterns from the target domain which results in a decreased training time compared to configuration1 where we only trained the fully-connected layer.

\section{Conclusion}
In this work, we introduce a system for recognizing human actions in theatre performances. The goal is to integrate this system into an output device that can provide descriptions of actors' actions on stage for individuals who are blind or visually impaired. To accomplish this, we collected a new dataset of theatre human actions using a Kinect sensor in a theatre environment. We then employed three spatio-temporal graph convolution networks (ST-GCN, 2s-AGCN, and MS-G3D) for the recognition task, utilizing transfer learning to achieve better performance and avoid the need for training from scratch, which would require a large number of training samples and significant computational resources.\\
To the best of our knowledge, this study is the first to investigate transfer learning on ST-GCN models for skeleton-based human action recognition. The results we obtained demonstrate the capability of these models to transfer previous knowledge, as well as learned spatial and temporal patterns from a source domain to new tasks. Additionally, we propose a framework for adapting the transfer learning process on the target domain based on its divergence from the source domain. The improvements we achieved validate the effectiveness of our proposed framework and highlight the importance of selecting appropriate transfer learning configurations based on the diversity and similarity between the source and target domains.\\
Ultimately, our efforts have successfully enhanced the recognition performance of pre-trained models on theatre actions.


\begin{thebibliography}{abbrv}

\bibitem{Ibel2022}
Farah Ibelaiden, Slimane Larabi.
Visual Place Representation and Recognition from Depth Images, Optik, Volume 260, June 2022, 169109

\bibitem{Zatout2021}
Chayma Zatout, Slimane Larabi.
Semantic scene synthesis: application to assistive systems . Vis Comput (2021). https://doi.org/10.1007/s00371-021-02147-w

\bibitem{Zatout2019}
Chayma Zatout, Slimane Larabi, Ilyes Mendili, Soedji Ablam Edoh Barnabe.
Ego-Semantic Labeling of Scene from Depth Image for Visually Impaired and Blind People.
ICCV 2019-EPIC, Seoul, November 2, 2019.

\bibitem{Zatout2020}
Chayma Zatout, Slimane Larabi.
A Novel Output Device for visually impaired and blind people's aid systems.
First International Conference on Communications, Control Systems and Signal Processing (CCSSP 2020), El Oued, Algeria, March 16-17, 2020. Link to the paper DOI: 10.1109/CCSSP49278.2020.9151820

\bibitem{Ibel2020}
Farah Ibelaiden, Brahim Sayah, Slimane Larabi.
Scene Descriptor from Depth Images for Visually Positioning.
First International Conference on Communications, Control Systems and Signal Processing (CCSSP 2020), El-Oued, Algeria, March 16-17, 2020. Link to the paper .DOI: 10.1109/CCSSP49278.2020.9151773.

\bibitem{Benhamida2022}
Leyla BenhamidaSlimane Larabi,
Human Action Recognition and Coding based on Skeleton Data for Visually Impaired and Blind People Aid System,
2022 First International Conference on Computer Communications and Intelligent Systems (I3CIS), November 2022

\bibitem{Delloul2022_2}
Delloul Khadidja, Slimane Larabi.
Egocentric Scene Description for the Blind and Visually Impaired .
5th International Symposium on Informatics and its Applications (ISIA), M’Sila University, November 29-30, 2022

\bibitem{Delloul2022}
Delloul Khadidja, Slimane Larabi.
Image Captioning State-of-the-Art: Is It Enough for the Guidance of Visually Impaired in an Environment? .
Advances in Computing Systems and Applications. 17-18 May, CSA 2022. Lecture Notes in Networks and Systems, vol 513. Springer, Cham.

\bibitem{Ibel2020_2}
Farah Ibelaiden, Slimane Larabi.
A Benchmark for Visual Positioning from Depth Images.
The 4th International Symposium on Informatics and its Applications (ISIA), Algeria, December, 15-16 2020.

\bibitem{1}
Shahroudy, Amir, et al. "Ntu rgb+ d: A large scale dataset for 3d human activity analysis." Proceedings of the IEEE conference on computer vision and pattern recognition. 2016.

\bibitem{2}
Liu, Jun, et al. "Ntu rgb+ d 120: A large-scale benchmark for 3d human activity understanding." IEEE transactions on pattern analysis and machine intelligence 42.10 (2019): 2684-2701.

\bibitem{3}
 Shi, Lei, et al. "Two-stream adaptive graph convolutional networks for skeleton-based action recognition." Proceedings of the IEEE/CVF conference on computer vision and pattern recognition. 2019.
 
 \bibitem{4}
Liao, Renjie, et al. "Lanczosnet: Multi-scale deep graph convolutional networks." arXiv preprint arXiv:1901.01484 (2019).

\bibitem{5}
Luan, Sitao, et al. "Break the ceiling: Stronger multi-scale deep graph convolutional networks." Advances in neural information processing systems 32 (2019).

\bibitem{6}
Wu, Zonghan, et al. "A comprehensive survey on graph neural networks." IEEE transactions on neural networks and learning systems 32.1 (2020): 4-24.

\bibitem{7}
Kipf, Thomas N., and Max Welling. "Semi-supervised classification with graph convolutional networks." arXiv preprint arXiv:1609.02907 (2016).

\bibitem{8}
Yan, Sijie, Yuanjun Xiong, and Dahua Lin. "Spatial temporal graph convolutional networks for skeleton-based action recognition." Proceedings of the AAAI conference on artificial intelligence. Vol. 32. No. 1. 2018.

\bibitem{9}
Fawaz, Hassan Ismail, et al. "Transfer learning for time series classification." 2018 IEEE international conference on big data (Big Data). IEEE, 2018.

\bibitem{10}
Taylor, Matthew E., and Peter Stone. "Transfer learning for reinforcement learning domains: A survey." Journal of Machine Learning Research 10.7 (2009).

\bibitem{11}
Li, Shuai, et al. "Independently recurrent neural network (indrnn): Building a longer and deeper rnn." Proceedings of the IEEE conference on computer vision and pattern recognition. 2018.

\bibitem{12}
Zhang, Chenyang, et al. "DAAL: Deep activation-based attribute learning for action recognition in depth videos." Computer Vision and Image Understanding 167 (2018): 37-49.

\bibitem{13}
Yang, Xiaodong, and YingLi Tian. "Super normal vector for activity recognition using depth sequences." Proceedings of the IEEE conference on computer vision and pattern recognition. 2014.

\bibitem{14}
Li, Maosen, et al. "Actional-structural graph convolutional networks for skeleton-based action recognition." Proceedings of the IEEE/CVF conference on computer vision and pattern recognition. 2019.

\bibitem{15}
Liu, Jun, et al. "Skeleton-based action recognition using spatio-temporal LSTM network with trust gates." IEEE transactions on pattern analysis and machine intelligence 40.12 (2017): 3007-3021.

\bibitem{16}
Li, Yanshan, et al. "Learning shape-motion representations from geometric algebra spatio-temporal model for skeleton-based action recognition." 2019 IEEE international conference on multimedia and Expo (ICME). IEEE, 2019.

\bibitem{17}
Xu, Yangyang, et al. "Ensemble one-dimensional convolution neural networks for skeleton-based action recognition." IEEE Signal Processing Letters 25.7 (2018): 1044-1048.

\bibitem{18}
Kim, Tae Soo, and Austin Reiter. "Interpretable 3d human action analysis with temporal convolutional networks." 2017 IEEE conference on computer vision and pattern recognition workshops (CVPRW). IEEE, 2017.

\bibitem{19}
Li, Shuai, et al. "Independently recurrent neural network (indrnn): Building a longer and deeper rnn." Proceedings of the IEEE conference on computer vision and pattern recognition. 2018.

\bibitem{20}
Li, Maosen, et al. "Actional-structural graph convolutional networks for skeleton-based action recognition." Proceedings of the IEEE/CVF conference on computer vision and pattern recognition. 2019.

\bibitem{21}
Abdulazeem, Yousry, et al. "Human action recognition based on transfer learning approach." IEEE Access 9 (2021): 82058-82069.

\bibitem{22}
Ray, Abhisek, et al. "Transfer Learning Enhanced Vision-based Human Activity Recognition: A Decade-long Analysis." International Journal of Information Management Data Insights 3.1 (2023): 100142.

\bibitem{23}
Wang, Jindong, et al. "Deep transfer learning for cross-domain activity recognition." proceedings of the 3rd International Conference on Crowd Science and Engineering. 2018.

\bibitem{24}
Kooverjee, Nishai, Steven James, and Terence Van Zyl. "Investigating transfer learning in graph neural networks." Electronics 11.8 (2022): 1202.

\bibitem{25}
Lacoste, Alexandre, et al. "Quantifying the carbon emissions of machine learning." arXiv preprint arXiv:1910.09700 (2019).

\bibitem{26}
Henderson, Peter, et al. "Towards the systematic reporting of the energy and carbon footprints of machine learning." The Journal of Machine Learning Research 21.1 (2020): 10039-10081.


\bibitem{28}
Cao, Zhe, et al. "OpenPose: realtime multi-person 2D pose estimation using Part Affinity Fields." IEEE transactions on pattern analysis and machine intelligence 43.1 (2021): 172-186.


\bibitem{29}
Shi, Lei, et al. "Skeleton-based action recognition with directed graph neural networks." Proceedings of the IEEE/CVF conference on computer vision and pattern recognition. 2019.

\bibitem{30}
Mukhiddinov, Mukhriddin, and Jinsoo Cho. "Smart glass system using deep learning for the blind and visually impaired." Electronics 10.22 (2021): 2756.

\bibitem{31}
Hegde, Pavan, et al. "Smart Glasses for Visually Disabled Person." International Journal of Research in Engineering and Science (IJRES) 9.7 (2021): 62-68.

\bibitem{32}
Kandalan, Roya Norouzi, and Kamesh Namuduri. "Techniques for constructing indoor navigation systems for the visually impaired: A review." IEEE Transactions on Human-Machine Systems 50.6 (2020): 492-506.

\bibitem{33}
Kumar, YR Sanjay, et al. "Smart glasses for visually impaired people with facial recognition." 2022 International Conference on Communication, Computing and Internet of Things (IC3IoT). IEEE, 2022.

\bibitem{34}
S. Bhole and A. Dhok, "Deep Learning based Object Detection and Recognition Framework for the Visually-Impaired," 2020 Fourth International Conference on Computing Methodologies and Communication (ICCMC), Erode, India, 2020, pp. 725-728.

\bibitem{35}
Liu, Ziyu, et al. "Disentangling and unifying graph convolutions for skeleton-based action recognition." Proceedings of the IEEE/CVF conference on computer vision and pattern recognition. 2020.



\end{thebibliography}
\end{document}